\documentclass{article}
\usepackage[preprint]{log_2022}
\usepackage{booktabs}            
\usepackage{multirow}            
\usepackage{amsfonts}            
\usepackage{graphicx}            
\usepackage{duckuments}          

\usepackage{times}  
\usepackage{adjustbox}
\usepackage{changepage}

\usepackage{tikz}
\usetikzlibrary{shapes, arrows, positioning}
\usepackage{adjustbox} 
\usepackage{helvet} 
\usepackage{courier}  
\usepackage{multirow}
\usepackage{graphicx} 
\usepackage{algorithm}
\usepackage[noend]{algpseudocode}
\usepackage{booktabs}
\usepackage{import}
\usepackage{amsmath}
\usepackage{amssymb}
\usepackage{subcaption}
\usepackage{todonotes}
\usepackage[toc,page,header]{appendix}
\usepackage{minitoc}

\usepackage{wrapfig}
\usepackage{amsmath}
\usepackage{mathtools}
\usepackage[bb=boondox,bbscaled=.95]{mathalfa}
\usepackage{tikz}
\usepackage{pgfplots}
\usepgfplotslibrary{groupplots}
\usepgfplotslibrary{patchplots}
\pgfplotsset{compat=1.14}

\usepackage{amsmath}
\usepackage{mathtools}
\usepackage[bb=boondox,bbscaled=.95]{mathalfa}
\usepackage{tikz}
\usepackage{pgfplots}
\usepgfplotslibrary{groupplots}
\usepgfplotslibrary{patchplots}
\pgfplotsset{compat=1.14}

\DeclareMathOperator{\R}{\mathbb{R}}

\DeclareMathOperator{\G}{\mathcal{G}}

\DeclareMathOperator{\Sa}{\mathcal{S}}

\newcommand{\Fb}{\mathbf{F}}
\newcommand{\Hb}{\mathbf{H}}

\newcommand{\Xb}{\mathbf{X}}

\usepackage[numbers,compress,sort]{natbib}

\title{Graph Neural Stochastic Differential Equations}

\author[R. Bergna]{Richard Bergna\\
\institute{University of Cambridge}\\
\email{rsb63@cam.ac.uk}\And
Felix L. Opolka \\
\institute{University of Cambridge}\\
\email{flo23@cam.ac.uk}\And
 Pietro Liò \\
\institute{University of Cambridge}\\
\email{pl219@cam.ac.uk}\And
Jose Miguel Hernandez-Lobato \\
\institute{University of Cambridge}\\
\email{jmh233@cam.ac.uk}
}

\begin{document}

\maketitle

\begin{abstract}

We present a novel model \textit{Graph Neural Stochastic Differential Equations} (Graph Neural SDEs). This technique enhances the \textit{Graph Neural Ordinary Differential Equations} (Graph Neural ODEs) by embedding randomness into data representation using Brownian motion. This inclusion allows for the assessment of prediction uncertainty, a crucial aspect frequently missed in current models. In our framework, we spotlight the \textit{Latent Graph Neural SDE} variant, demonstrating its effectiveness. Through empirical studies, we find that Latent Graph Neural SDEs surpass conventional models like Graph Convolutional Networks and Graph Neural ODEs, especially in confidence prediction, making them superior in handling out-of-distribution detection across both static and spatio-temporal contexts.
\end{abstract}

\section{Introduction}

Before the widespread use of neural networks and modern machine learning, differential equations, including Stochastic Differential Equations (SDEs) were the gold standard for modeling systems across diverse scientific disciplines \citep{hoops2016ordinary, quach2007estimating, mandelzweig2001quasilinearization, cardelli2008processes, buckdahn2011general, cvijovic2014bridging}. In the machine learning arena, neural  networks' integration with ODEs has enabled advanced continuous-time data modeling \cite{chen2018neural, li2020scalable}. Despite these progresses, the fusion of SDEs and Graph Neural Networks (GNNs) is still unexplored. 

This work presents the \textbf{Graph Neural SDE}, a model that harnesses the robustness of SDEs and the versatility of GNNs to handle complex graph-structured data. Due to its stochastic nature, this model also enables precise uncertainty quantification in its predictions. Furthermore, we uncover a deep theoretical connection between our Graph Neural SDEs and the continuous representations of deep Graph Recurrent Neural Networks (refer to Appendix \ref{A:RNN_REsNET}). Additionally, we highlight parallels between Graph Neural ODEs and continuous deep Graph Residual Neural Networks, mirroring findings by \cite{li2020scalable} and \cite{chen2018neural}.

Of particular interest, we compare the Graph Neural SDE against existing uncertainty quantification techniques for GNNs, notably the Bayesian GNN \citep{hasanzadeh2020bayesian} and Ensemble of GNN methods \citep{lin2022robust} - two of the few established approaches for uncertainty quantification in graph data, as documented in the literature. Our experiments, shows superior performance of the Graph Neural SDE, which in many datasets surpasses both Bayesian and Ensemble approaches. Furthermore, our experiments show the Graph Neural SDE out perform Graph Neural ODEs in most spectrum of tasks, including both static and spatio-temporal datasets. The paper further details the formulation and implementation of our models, provides intuitive visualizations, and validates our approach using real-world datasets.

\section{Background}

\subsection{Neural Ordinary Differential Equations}
Neural ODEs provide an elegant approach to modeling dynamical systems using the principles of neural networks. Instead of representing data transformations as discrete layers in a traditional deep network, Neural ODEs describe them as continuous transformations parameterized by differential equations. This concept of continuous transformations within the neural network is often termed as "continuous-depth", implying that instead of having distinct layers, the network smoothly transitions and evolves data through a continuum of depths.  

These transformations specify how the state of a system, denoted as \( z(t) \), evolves over time. The rate of change in the state of the system is determined by a function \( f \), which is parameterized by neural network weights. This relationship is represented as

\begin{equation*}
\begin{minipage}{0.2\linewidth}
\centering
$$\frac{dz}{dt} = f(z(t), t),$$
\end{minipage}%
\begin{minipage}{0.2\linewidth}
\centering
$$z(0) = z_0.$$
\end{minipage}
\end{equation*}

This differential equation implies that the state of the system at any time \( t \) is determined by accumulating the effects of the function \( f \) from the initial state \( z_0 \) up to that time. This can be articulated more explicitly as

\begin{equation*}
z_t = z_0 + \int_{0}^{t} f(z(s), s) \, \text{d}s\footnote{In the integral expression, \( s \) is a dummy variable of integration, representing the intermediate points between 0 and \( t \) over which the function \( f \) is integrated.}.
\end{equation*}

In the Neural ODE framework, the unknown function \( f \), governing system dynamics, is approximated using a neural network.

\subsection{Neural Stochastic Differential Equations}
Stochastic Differential Equations (SDEs) have been widely employed to model real-world phenomena that exhibit randomness, such as physical systems, financial markets, population dynamics, and genetic variations \citep{schreiber2011persistence, gontis2014consentaneous, ricciardi2012biomathematics}. They generalize ODEs by modeling systems that evolve continuously over time, incorporating randomness. Informally, an SDE can be seen as an ODE that integrates a certain degree of noise

\begin{equation*}
    \frac{\text{d}z}{\text{d}t} = f(z(t), t) + \epsilon(t).
\end{equation*}

Here, $\epsilon(t)$ represents the time-dependent noise, typically modeled using diffusion models and Brownian motion. In a more formal definition, an SDE is

\begin{equation*}
\text{d}z(t) = \underbrace{f(t, z(t))}_{\text{drift}} \text{d}t + \underbrace{g(t, z(t))}_{\text{diffusion}} \text{d}W(t).
\end{equation*}

In this equation, the system state \(z(t)\) at time \(t\) evolves due to two main components: the drift function \(f\) and the diffusion function \(g\). The term \(\text{d}W(t)\) denotes the infinitesimal increment of a standard Brownian motion (or Wiener process) \(W(t)\), with properties like \(W(0) = 0\), independent increments, and \(W(t) - W(\tau)\) being normally distributed with mean 0 and variance \(t-\tau\) for \(0 \leq \tau < t\). The strong solution for the SDE, denoted as \(z(t)\), exists and is unique under conditions where \(f\) and \(g\) are Lipschitz and \(E[z(0)^2] < \infty\)\footnote{For a comprehensive and rigorous exploration of Stochastic Differential Equations, readers are referred to \cite{khasminskii2012stochastic} and \cite{revuz2013continuous}.}.

The \textbf{drift function} \(f(t, z(t))\) represents the deterministic component of the system's evolution, describing the expected direction of change at each time point based on the current state \(z(t)\). 

The \textbf{diffusion function} \(g(t, z(t))\) characterizes the system's random component, scaling the random noise introduced by the Wiener process \(W(t)\), commonly known as Brownian motion.

Within the Neural SDE framework, analogous to Neural ODEs, the SDE \citep{li2020scalable} is numerically approximated. This involves evaluating the system's response to both deterministic and random effects over time. The solution to the given SDE is represented as

\begin{equation*}
z(t) = z(0) + \int_{0}^{t} f(s, z(s)) \text{d}s + \int_{0}^{t} g(s, z(s)) \text{d}W(s).
\end{equation*}

This equation provides an integral expression of how the state \( z(t) \) evolves, subject to both deterministic (through the function \( f \)) and random influences (through the function \( g \) and the Wiener process \( W(t) \)).

One challenge in Neural SDEs is when the diffusion function \(g\) becomes a learnable parameter and is trained to achieve maximum likelihood, such as by directly minimizing cross-entropy or mean square error. In these situations, the diffusion function often converges to zero, turning the Neural SDE into a Neural ODE. To address this, researchers have suggested strategies like minimizing the Kullback-Leibler (KL) divergence or Wasserstein distance, forming the foundation for advanced concepts like `Latent SDEs' and `SDE-Generative Adversarial Networks' (SDE-GANs) \citep{li2020scalable, kidger2021efficient}.

\subsection{Graph Neural Ordinary Differential Equations}

Introduced by \citet{poli2019graph}, Graph Neural Ordinary Differential Equations (GN-ODEs) combine
continuous-depth adaptability from the Neural ODEs with graph neural network structure. GN-ODEs meld the structured representation of graph data with the continuous model flexibility, providing a continuum of GNN layers. Compatible with both static and autoregressive GNN models, GDEs afford computational advantages in static contexts using the adjoint method \citep{chen2018neural} and enhance performance in dynamic situations by leveraging the geometry of the underlying dynamics.

At the heart of GN-ODEs is the representation of the dynamics between layers of GNN node feature matrices
\begin{equation*}
\begin{minipage}{0.4\linewidth}
\centering
        $$\Hb(s+1) = \Hb(s) + \Fb_{\G}(s, \Hb(s), \theta(s)),$$
\end{minipage}%
\begin{minipage}{0.2\linewidth}
\centering
        $$\Hb(0) = \Xb_e.$$
\end{minipage}
\end{equation*}

In this representation, \( \Xb \) denotes the initial node features of the graph, and \( \Xb_e \) is an embedding derived from various methods such as a single linear layer or another GNN layer. The function \( \Fb_{\G} \) represents a matrix-valued nonlinear function conditioned on graph \( \mathcal{G} \), and \( \theta(s) \) is the tensor of parameters for the \( s \)-th layer. The GN-ODE model can be more succinctly expressed as:
\begin{equation*}
\begin{minipage}{0.4\linewidth}
\centering
    $$\dot\Hb(s) = \Fb_{\G}(s, \Hb(s), \theta),$$
\end{minipage}%
\begin{minipage}{0.2\linewidth}
\centering
    $$\Hb(0) = \Xb_e,$$
\end{minipage}
\end{equation*}
where \( s \) belongs to a subset \( \Sa \) of the real numbers, \( \R \), typically denoted as \([t_0, t_1]\).

In the context of GN-ODEs, \( \Fb_{\G} \) functions as a field on graph \( \mathcal{G} \) that varies with the depth or complexity of the model, which we refer to as "depth-varying". Depending on the context, the node features might be augmented to improve both computational efficiency and the model's performance, as indicated in prior studies.

\section{Graph Neural SDEs}
Drawing inspiration from Graph Neural ODEs \citep{poli2019graph} and Latent SDEs \citep{li2020scalable}, we introduce our methodology: Latent Graph Neural SDEs, leveraging the latent strategy for Neural SDEs.

\subsection{Latent Graph Neural SDEs}

Latent Graph Neural SDEs learn an latent state \( z(t) \) using Graph Neural SDEs to encapsulate the data's underlying patterns. Once determined, this state is input into a projection network \( f_{\Omega} \) to generate predictions \( \hat{y} \). The model parameterizes an Ornstein–Uhlenbeck (OU) prior process and an approximate posterior, which is another OU process.

More formally, the prior is defined by
\[
\text{d}\tilde{z}(t) = f_{\theta}(z(t), t,\mathcal{G})\text{d}t + \sigma(\tilde{z}_t, t) dW_t,
\]
where \( f_{\theta} \) is typically set to a constant (e.g., 0 in our experiments) and \( \sigma \) is set to a fixed value, such as 1.0. 

The approximate posterior is
\[
\text{d}z(t) =f_{\phi}(z(t), t,\mathcal{G})\text{d}t + \sigma(z_t, t) dW_t.
\]
Here, \( f_{\phi} \) is parameterized by a neural network, with \( \phi \) representing the learned weights of the network.

Both the prior and posterior drift functions, \( f_{\theta} \) and \( f_{\phi} \) respectively, dictate the dynamics of the system. \( z(t) \) denotes the system state at time \( t \), and \( \mathcal{G} \) symbolizes the graph structure. Notably, both the prior and posterior SDEs employ the same diffusion function \( \sigma \) but have distinct drift functions. Sharing the diffusion function ensures that the KL divergence between these processes remains finite, facilitating its estimation by sampling paths from the approximate posterior process \citep{li2020scalable}. The KL divergence between these processes is finite and can be estimated by sampling paths from the approximate posterior process.

The evidence lower bound (ELBO) is given by
\[
\mathcal{L}_{ELBO}(\phi) =  \mathbb{E}_{z_{t}} \left[\log(p(x_{t_i}|z_{t_i}) - \int_{t_0}^{t_1} \frac{1}{2} ||u(z_t, t, \phi, \theta, \mathcal{G})||^2_2 \text{d}t\right],
\]
where \( x_{t_i} \) are the observations at time \( t \) (with \( i \) in \([t_0, t_1]\)), and
\[
u = g(z_t, t)^{-1}[f_{\phi}(z_t, t, \mathcal{G}) - f_{\theta}(z_t, t, \mathcal{G})].
\]

Given the latent state \( z_t \), it's fed into the projection layer, \( f_{\Omega} \), for further prediction. The posterior predictive is then

\[
p(y^{*}|t^{*}, \mathcal{G}, \mathcal{D}) = \int p \left(y^{*}|f_{\Omega}(z_{t},t^{*}, \mathcal{G}) \right) p(z|\mathcal{D}) \, \text{d}z \approx \frac{1}{N} \sum_{n=1}^{N} p \left(y^{*}|f_{\Omega}(z_n,t^{*}, \mathcal{G}) \right).
\]

As shown on the right side of this equation, the predictive distribution is approximated using Monte Carlo sampling by drawing samples \( z_n \) from the posterior \( p(z|\mathcal{D}) \). The variance, of the Monte Carlo mean estimation, is given by

\[
\text{Var}(y) = \frac{1}{N} \sum_{n=1}^{N} (y_n - \bar{y})^2.
\]

Furthermore, this Graph Neural SDE can be mathematically related to popular Graph Neural Networks, like Graph RNNs. For a comprehensive insight into the relationship between Graph Neural ODEs, Graph Neural SDEs, and continuous deep Residual and Recurrent Graph Neural Networks, please refer to Appendix \ref{A:RNN_REsNET}.

\subsection{Static Dataset}

\begin{figure}[ht]
  \centering

    \begin{subfigure}{0.40\textwidth}
    \centering
    \includegraphics[width=\linewidth]{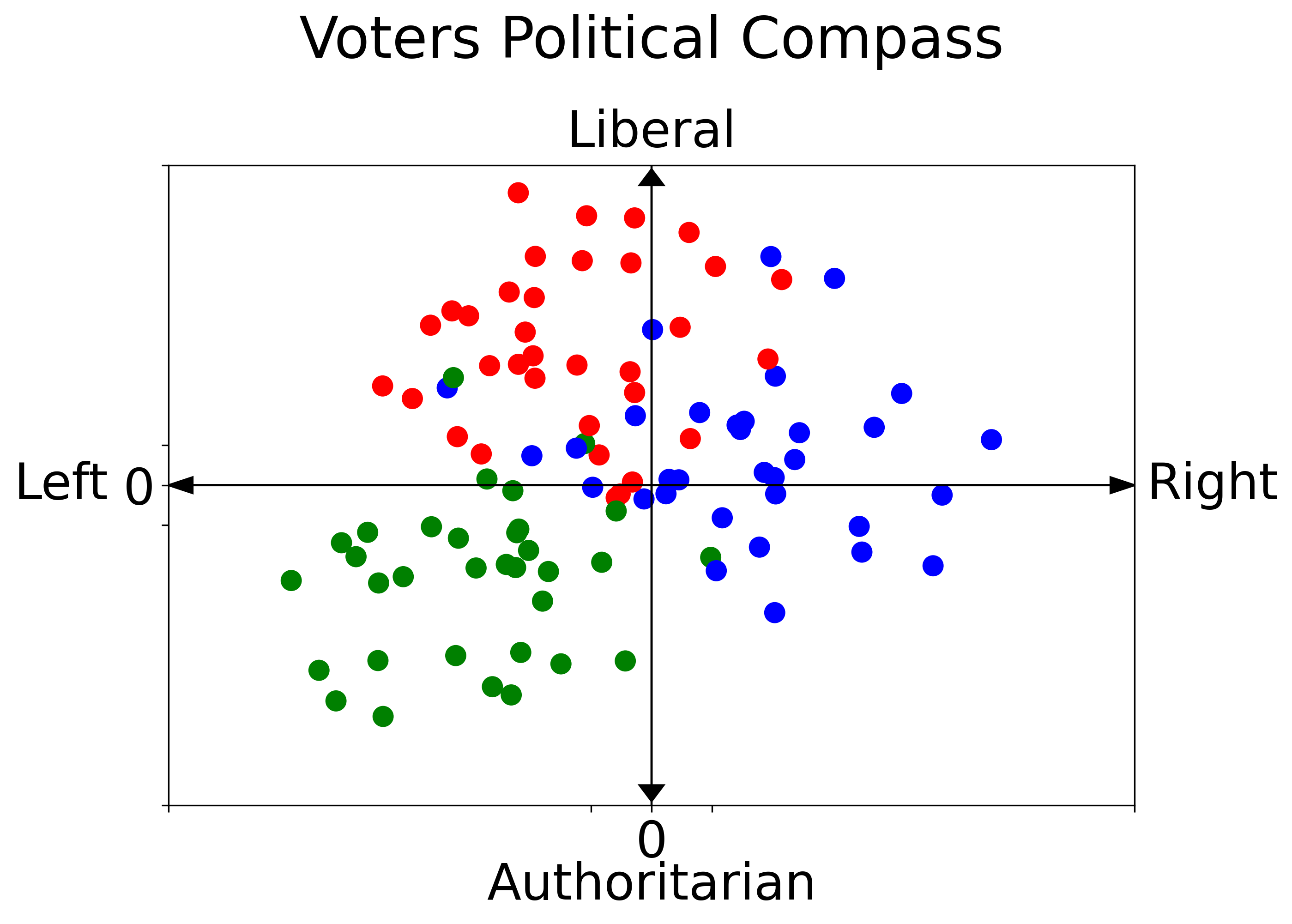}

  \end{subfigure}%
  \begin{subfigure}{0.38\textwidth}
    \centering
    \includegraphics[width=\linewidth]{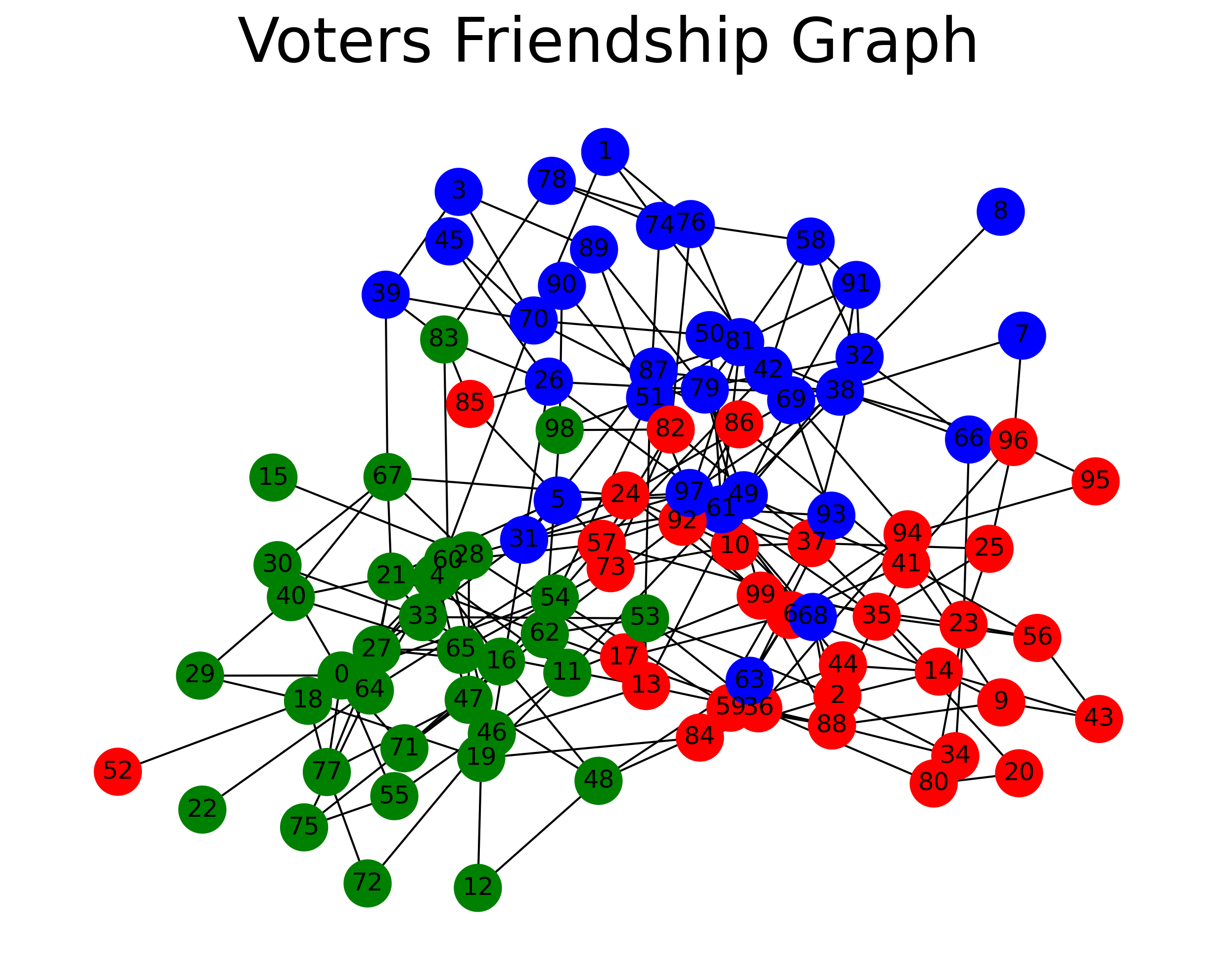}
  \end{subfigure}
  \caption{The left image illustrates the political compass of voters while the right image presents their social circles, with colors indicating the candidates they voted for.}
\label{fig:toy_dataset_problem}

\end{figure}

We aim to predict individual voting preferences for three candidates based on their political compass and social circles, as depicted in Figure \ref{fig:toy_dataset_problem}. The data contains inherent noise, with individuals maintaining friendships across voting preferences and their political compass not strictly dictating their voting choice, introducing randomness.

\begin{figure}[ht]
  \centering

    \begin{subfigure}{0.48\textwidth}
    \centering
    \includegraphics[width=\linewidth]{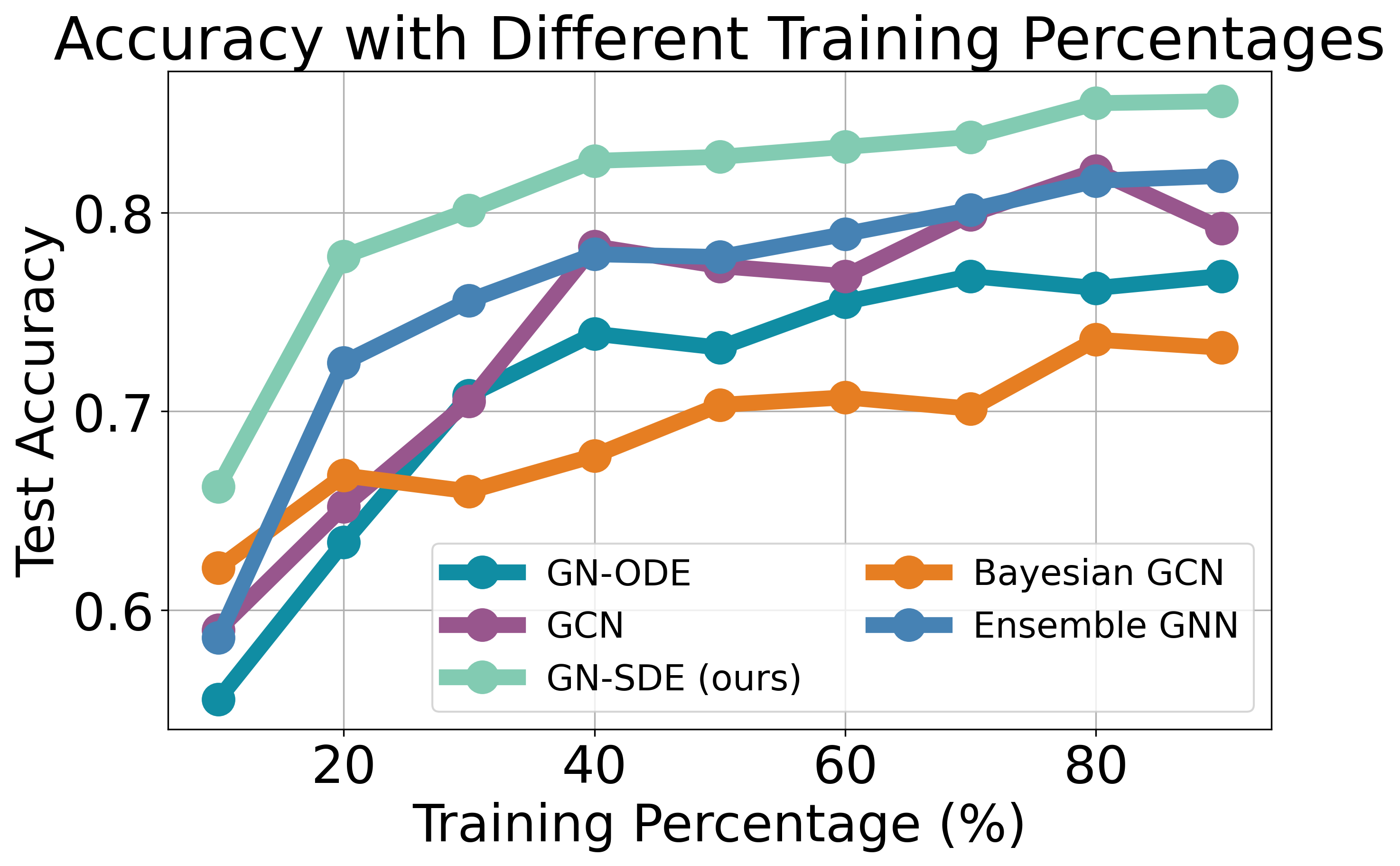}
  \end{subfigure}%
  \begin{subfigure}{0.45\textwidth}
    \centering
    \includegraphics[width=\linewidth]{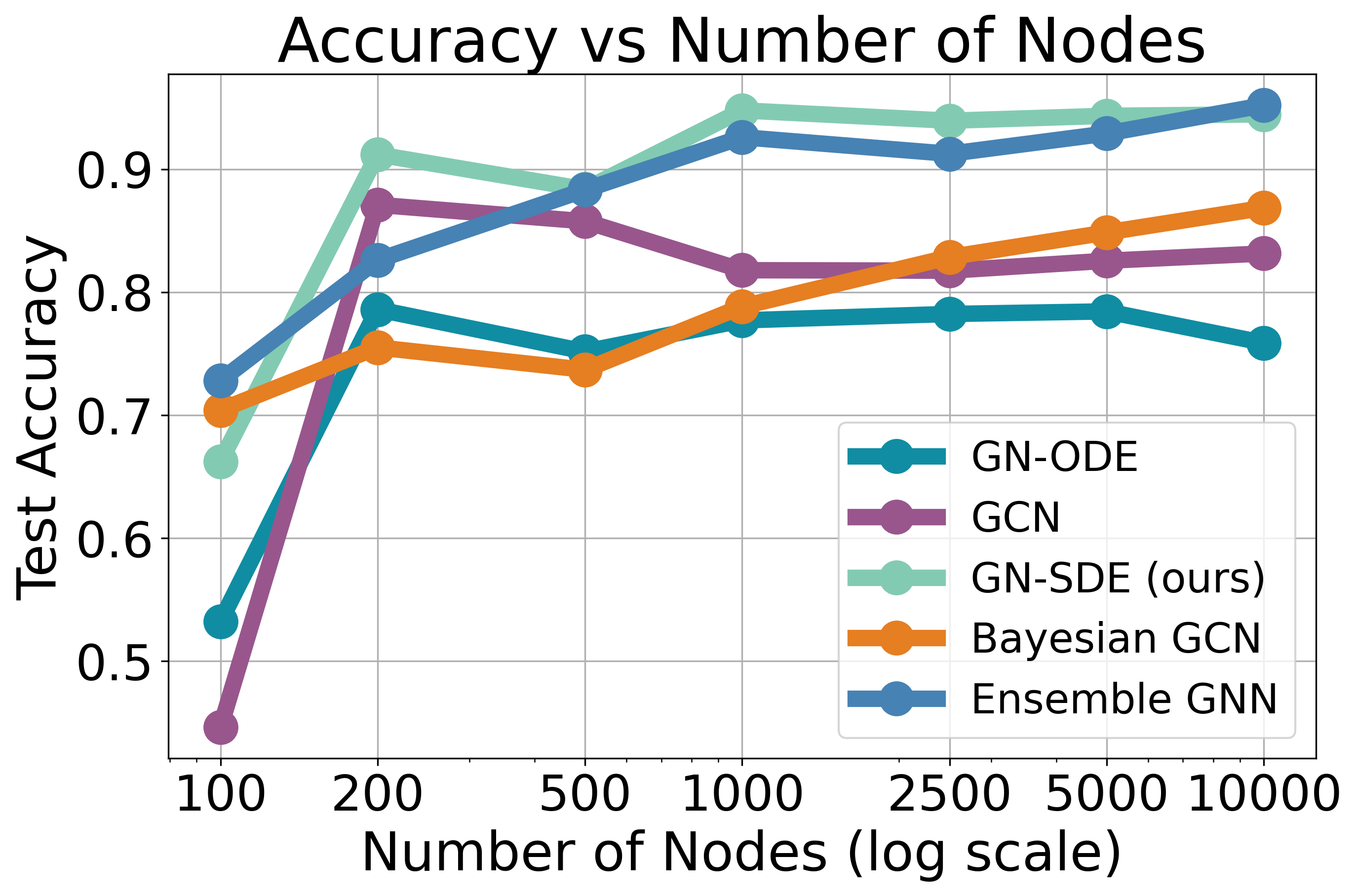}
  \end{subfigure}

    \begin{subfigure}{0.48\textwidth}
    \centering
    \includegraphics[width=\linewidth]{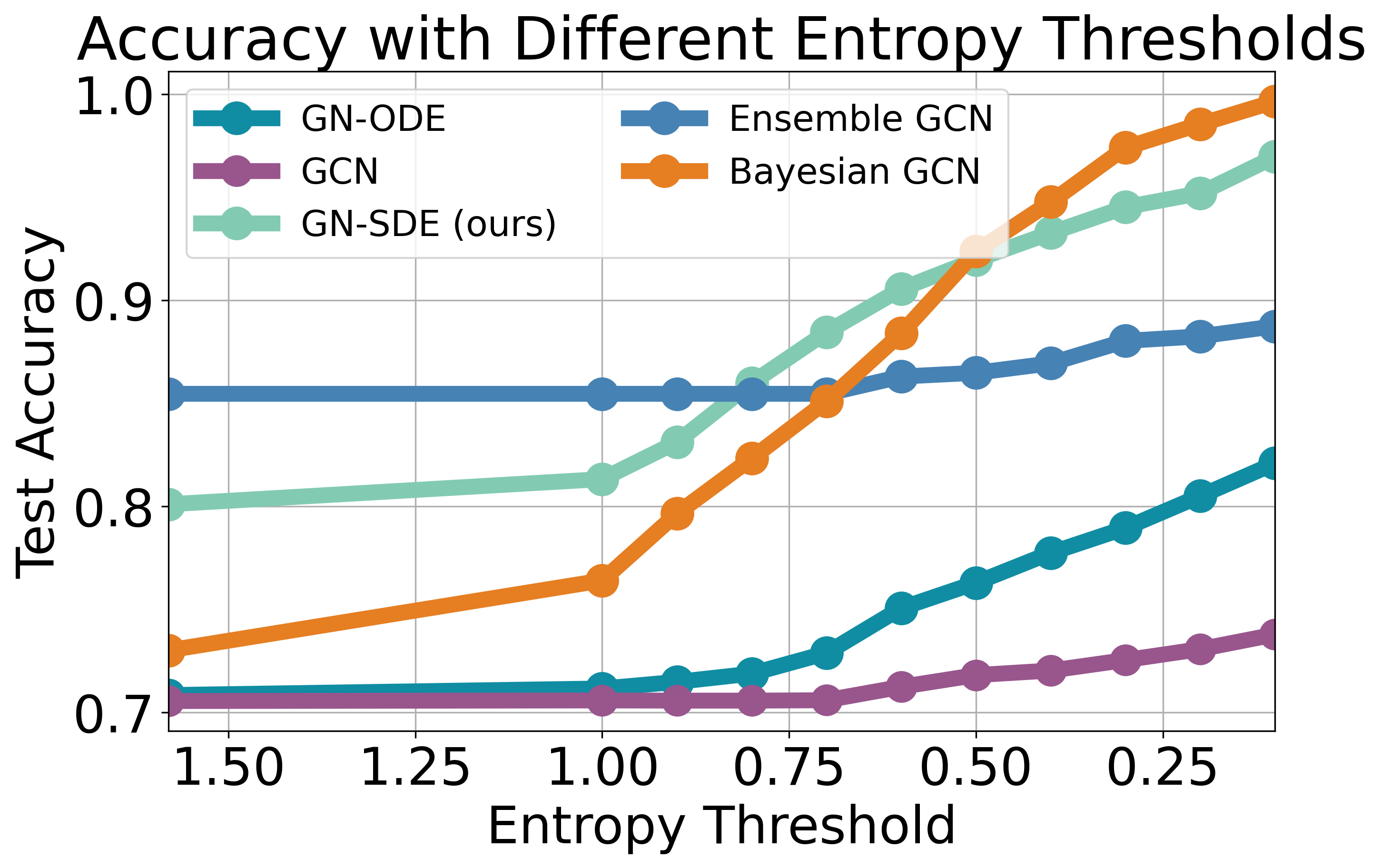}
  \end{subfigure}
    \hspace{0.02\textwidth}
  \begin{subfigure}{0.45\textwidth}
    \centering
    \includegraphics[width=\linewidth]{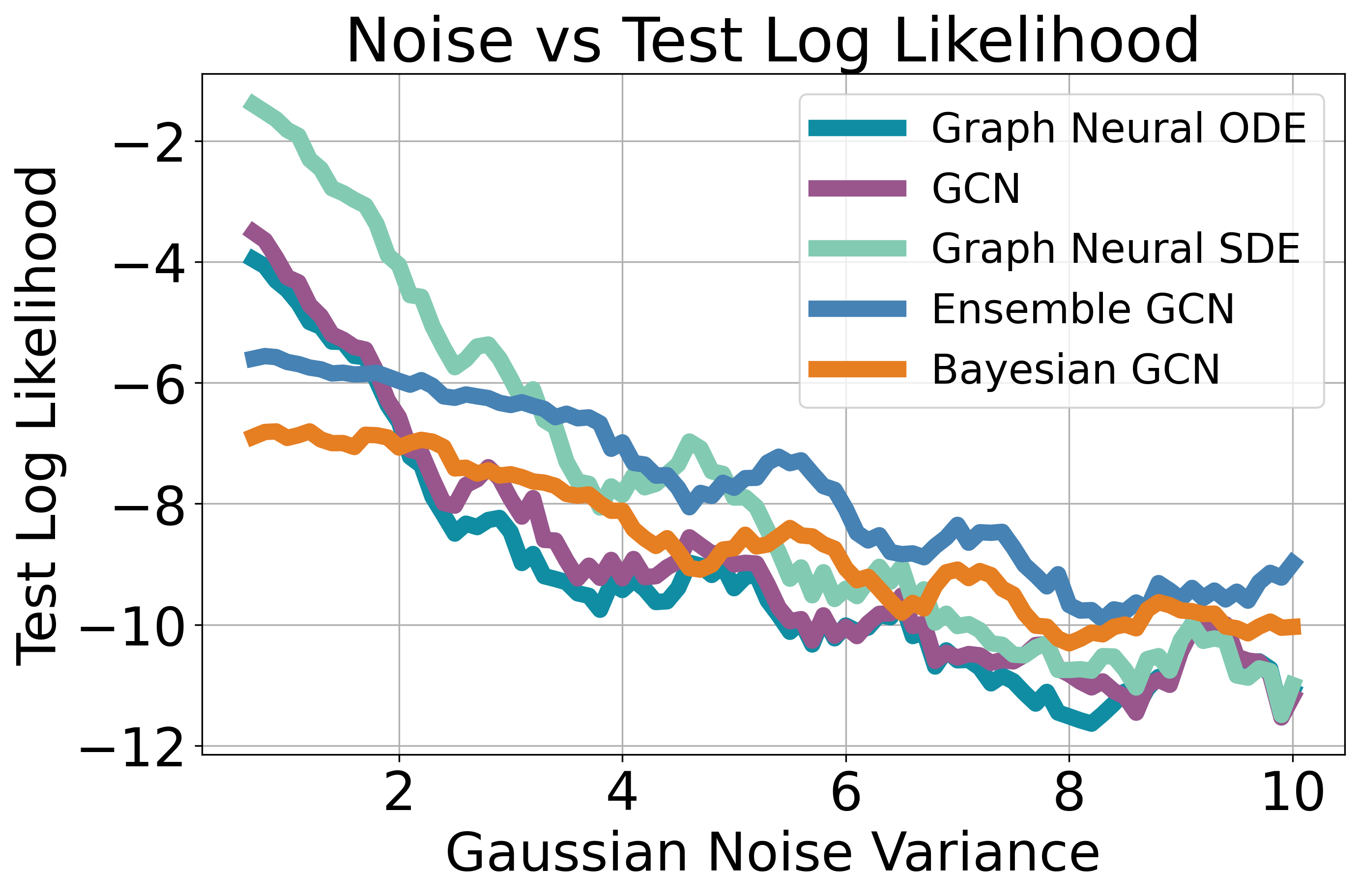}
  \end{subfigure}%
  \caption{Comparative evaluation of Graph Neural ODE, GCN, and Graph Neural SDE models.}
  \label{toy_dataset_four_imagess}
\end{figure}

Figure \ref{toy_dataset_four_imagess} presents a detailed comparison across five models: GN-ODE, GCN, our Latent GN-SDE, Bayesian GCN, and Ensemble GCN — with the last one averaging predictions from five individual GCNs. The evaluation is segmented into four metrics, each depicted in its respective sub-figure.

\textbf{Accuracy vs. Training Data Proportions:} Training on dataset portions from 10\% to 90\%, the GN-SDE consistently outperformed the other models, highlighting its data efficiency.

\textbf{Accuracy vs. Number of Nodes:} The GN-SDE maintained top performance across varying node counts, showcasing its adaptability to both small and large graphs. At 100 nodes, the Ensemble GCN momentarily exceeded GN-SDE, but the latter remained dominant in most scenarios.

\textbf{Accuracy vs. Entropy Threshold:} The entropy threshold is inversely related to the model's confidence in its predictions; the lower the entropy, the higher the confidence required for a prediction. In this context, our model demonstrated exemplary performance in identifying out-of-distribution data and providing accurate measures of uncertainty. While our model surpassed the performance of the Ensemble GCN, it was slightly outperformed by the Bayesian GCN at very low entropy thresholds, still indicating high confidence and quality in uncertainty quantification.

\textbf{Noise vs. Log-Likelihood:} Evaluating resilience to added Gaussian noise, the GN-SDE model exhibited the most gradual performance decline compared to competitors like the GCN and GN-ODE. Bayesian and Ensemble models performed slightly better under high noise.

To conclude, the GN-SDE consistently surpassed GN-ODE and GCN in every test. While Bayesian GCN and the Ensemble model occasionally outperformed ours, the GN-SDE stood out in robustness, data efficiency, and uncertainty quantification.

\begin{figure}[ht]
  \centering
    
  \begin{subfigure}[b]{0.45\textwidth}
    \includegraphics[width=\linewidth]{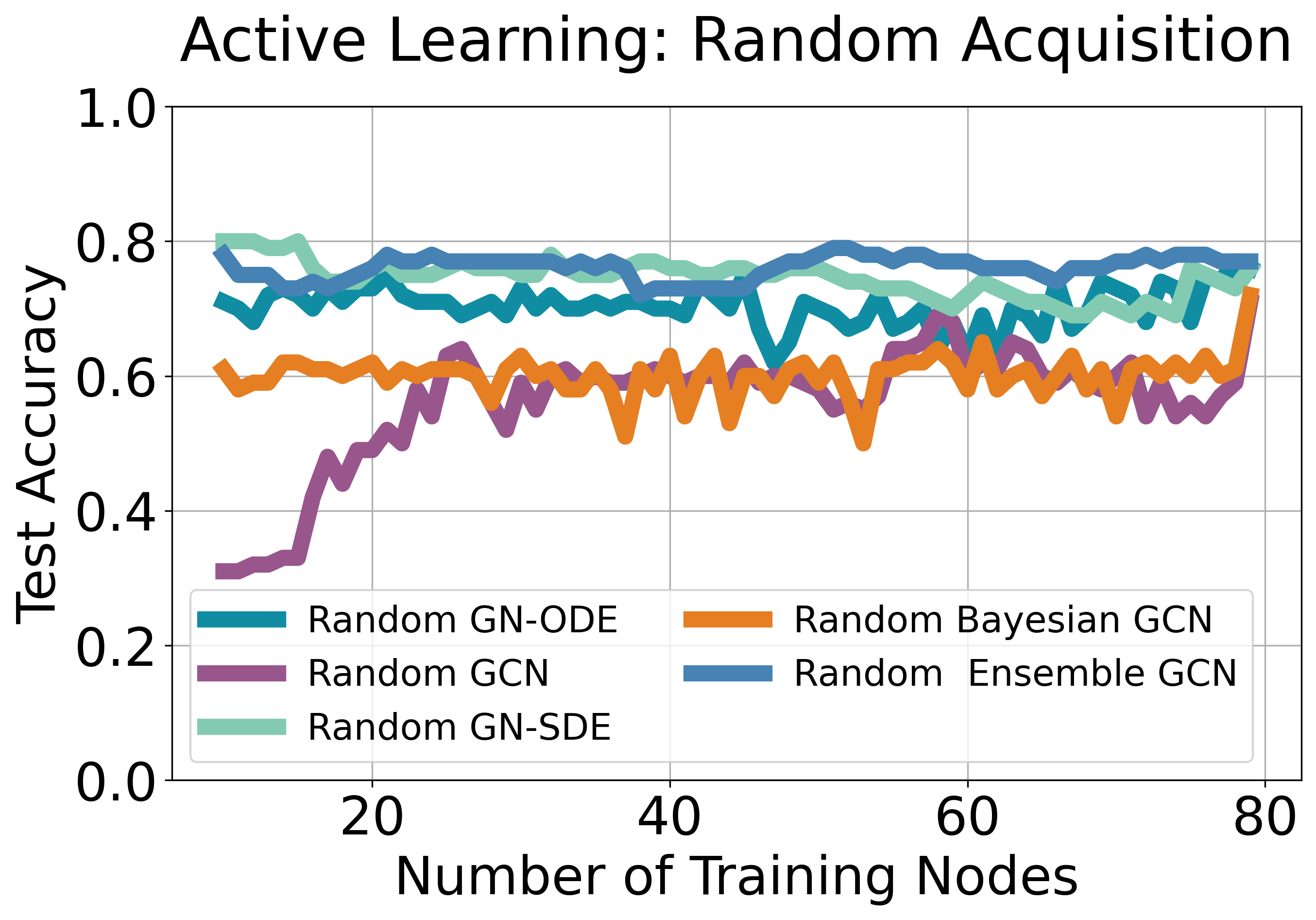}
    \caption{Node selection using a random acquisition function.}
  \end{subfigure}
  \hfill 
  \begin{subfigure}[b]{0.45\textwidth}
    \includegraphics[width=\linewidth, height=4.35cm]{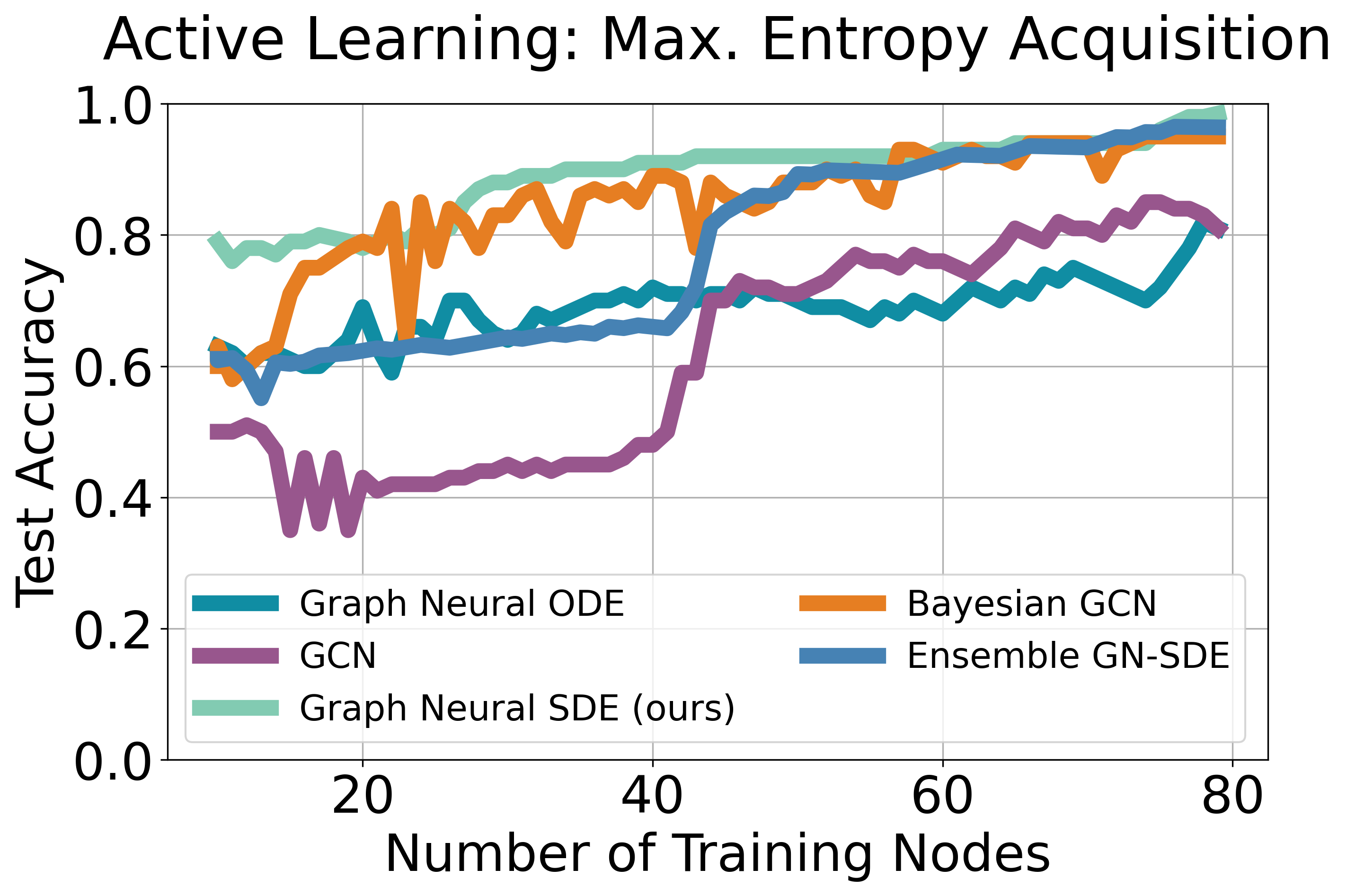}
    \caption{Node selection using a max entropy acquisition function.}
  \end{subfigure}
  \caption{The figure depicts an active learning experiment on a 100-node dataset, starting with 10 nodes and incrementally adding more until reaching 80. The left and right figures use random and max entropy acquisition functions respectively for node selection.}
  \label{fig:toy_dataset_problem_active_learning}
\end{figure}

Figure \ref{fig:toy_dataset_problem_active_learning} depicts an active learning study on a 100-node dataset. Starting with 10 nodes, additional nodes were sequentially added: based on the highest entropy in the right figure, and randomly in the left. After each addition, the model was trained over 5 epochs, continuing until 80 nodes were included. Our model reached a notable 99\% accuracy at the 78th node, closely matched by the Bayesian and Ensemble models. In comparison, GCN and Graph Neural ODE peaked around 85\%. This demonstrates our model's strength in active learning, especially vital when labeled data is limited or expensive. The random method's ceiling at 80\% accuracy highlights the benefit of uncertainty-aware strategies.

\subsubsection{Real-Word Static Data sets}
\begin{table}[ht]
    \centering
    \begin{adjustbox}{width=\textwidth}
    \begin{tabular}{l l cccccccccccc}
    \toprule
    \multirow{2}{*}{\underline{Dataset}} & \multirow{2}{*}{\underline{Models}} & \multicolumn{12}{c}{Entropy Thresholds} \\
    \cmidrule(r){3-14}
    \multicolumn{1}{c}{} & & $\infty$ & 1.6 & 1.0 & 0.9 & 0.8 & 0.7 & 0.6 & 0.5 & 0.4 & 0.3 & 0.2 & 0.1 \\ \midrule
    \multicolumn{1}{c}{} & \textbf{GN-SDE (ours)} & \textbf{0.817} & \textbf{0.834} & 0.922 & 0.942 & \textbf{0.957} & 0.966 & 0.969 & 0.977 & \textbf{0.991} & \textbf{\textbf{1.0}} & \textbf{\textbf{1.0}} & \textbf{\textbf{1.0}} \\
    \multicolumn{1}{c}{} & GN-ODE & 0.799 & 0.806 & 0.905 & 0.923 & 0.944 & \textbf{0.969} & \textbf{0.976} & \textbf{0.984} & 0.984 & 0.995 & \textbf{1.0} & \textbf{1.0} \\
    \multicolumn{1}{c}{CORA} & GCN & 0.717 & 0.717 & 0.720 & 0.720 & 0.723 & 0.734 & 0.756 & 0.761 & 0.771 & 0.780 & 0.786 & 0.824 \\ 
    \multicolumn{1}{c}{} & Ensemble GNN & 0.777 & 0.802 & \textbf{0.935} & \textbf{0.949} & 0.954 & 0.958 & 0.962 & 0.972 & 0.983 & \textbf{1.0} & \textbf{1.0} & \textbf{1.0} \\
    \multicolumn{1}{c}{} & Bayesian GNN & 0.709 & 0.719 & 0.800 & 0.834 & 0.871 & 0.893 & 0.917 & 0.925 & 0.930 & 0.948 & 0.972 & 0.981 \\ \midrule
    \multicolumn{1}{c}{} & \textbf{GN-SDE (ours)} & 0.71 & \textbf{0.753} & \textbf{0.879} & 0.889 & 0.898 & 0.925 & \textbf{0.929} & 0.924       & 0.926 & \textbf{0.947} & \textbf{0.972} & \textbf{1.0} \\
    & GN-ODE & \textbf{0.712} & 0.742 & 0.875 & \textbf{0.891} & \textbf{0.905} & \textbf{0.931} & 0.915 & \textbf{0.936} & \textbf{0.930} & 0.882 & 0.923 & \textbf{1.0} \\
    Citeseer & GCN & 0.516 & 0.516 & 0.524 & 0.530 & 0.544 & 0.557 & 0.583 & 0.599 & 0.626 & 0.651 & 0.678 & 0.725 \\
    & Bayesian GCN & 0.61 & 0.619 & 0.729 & 0.746 & 0.769 & 0.791 & 0.815 & 0.821 & 0.854 & 0.863 & 0.867 & 0.88 \\
    & Ensemble GNN & 0.527 & 0.532 & 0.743 & 0.780 & 0.821 & 0.834 & 0.859 & 0.858 & 0.920 & 0.939 & 0.953 & 0.933 \\ \midrule
   \multicolumn{1}{c}{} & \textbf{GN-SDE (ours)} & \textbf{0.791} & \textbf{0.791} & \textbf{0.794} & 0.794 & 0.802 & 0.818 & 0.837 & 0.852 & 0.876 & 0.893 & 0.898 & 0.911 \\
    \multicolumn{1}{c}{} & GN-ODE & 0.763 & 0.763 & 0.768 & 0.774 & 0.781 & 0.813 & 0.833 & 0.847 & 0.858 & 0.862 & 0.872 & 0.899 \\
    \multicolumn{1}{c}{Pubmed} & GCN & 0.78 & 0.78 & 0.784 & 0.785 & 0.789 & 0.795 & 0.809 & 0.821 & 0.823 & 0.835 & 0.847 & 0.864 \\ 
    \multicolumn{1}{c}{} & Ensemble GNN & 0.786 & 0.786 & 0.796 & \textbf{0.814} & \textbf{0.837} & \textbf{0.868} & \textbf{0.879} & \textbf{0.908} & \textbf{0.907} & \textbf{0.916} & \textbf{0.929} & \textbf{0.952} \\
    \multicolumn{1}{c}{} & Bayesian GNN & 0.715 & 0.715 & 0.719 & 0.730 & 0.736 & 0.752 & 0.815 & 0.850 & 0.864 & 0.882 & 0.904 & 0.918 \\
    \midrule
    \multicolumn{1}{c}{} & \textbf{GN-SDE (ours)} & \textbf{0.531} & 0.770 & 0.859 & 0.871 & 0.884 & 0.897 & 0.907 & 0.916 & 0.929 & 0.944 & 0.955 & 0.968 \\
    \multicolumn{1}{c}{} & GN-ODE & 0.526 & 0.766 & 0.853 & 0.867 & 0.883 & 0.898 & 0.912 & 0.919 & 0.930 & 0.938 & 0.951 & 0.964 \\
    \multicolumn{1}{c}{OGB arXiv} & GCN & 0.470 & \textbf{0.809} & \textbf{0.885} & \textbf{0.900} & 0.909 & 0.917 & 0.931 & \textbf{0.939} & \textbf{0.950} & \textbf{0.966} & \textbf{0.983} & 0.976 \\
    \multicolumn{1}{c}{} & Ensemble GNN & 0.512 & 0.785 & 0.699 & 0.800 & \textbf{1.000} & \textbf{1.000} & \textbf{1.000} & - & - & - & - & - \\
    \multicolumn{1}{c}{} & Bayesian GNN & 0.433 & 0.828 & 0.880 & 0.884 & 0.893 & 0.897 & 0.900 & 0.919 & 0.916 & 0.935 & 0.954 & \textbf{1.000} \\

    \bottomrule
    \end{tabular}
    \end{adjustbox}
    \caption{Accuracy scores for GN-SDE, GN-ODE, GCN, Ensemble GNN, and Bayesian GNN on CORA, Citeseer, Pubmed, and OGB arXiv datasets. Comparisons use varying entropy thresholds, with bold values indicating top performance. A '-' for accuracy indicates the model lacked sufficiently confident data points at that threshold.}

    \label{static_real_world_dataset}
\end{table}

Shifting our focus to real-world datasets, Table \ref{static_real_world_dataset} details the performance of the models GN-SDE, GN-ODE, GCN, Ensemble GCN, and Bayesian GCN on renowned datasets such as CORA \citep{sen2008collective}, Pubmed \citep{sen2008collective}, Citeseer \citep{giles1998citeseer}, and OGB arXiv  \citep{hu2020open}. The models were evaluated based on different entropy thresholds. This essentially means that a model is only permitted to make predictions if its confidence level, as measured by entropy, falls below a predefined threshold. Using entropy as a measure provides an understanding of a model's capacity for confident predictions (uncertainty quantification) and its ability in performing out-of-distribution detection.

A review of the table shows our GN-SDE model excelling across most entropy thresholds, particularly in the CORA and Pubmed datasets. In CORA, GN-SDE achieves 100\% accuracy at thresholds 0.3, 0.2, and 0.1, outperforming or matching other models. In Pubmed, GN-SDE leads across most thresholds, peaking at 91.1\% accuracy for a 0.1 entropy threshold.

For Citeseer and OGB arXiv, GN-SDE's performance is more competitive. In OGB arXiv, while GN-SDE starts strong, the Bayesian GNN reaches 100\% accuracy at a 0.1 entropy threshold. Ensemble GNN also impresses, hitting 100\% accuracy for thresholds up to 0.8. In Citeseer, GN-SDE often leads, with GN-ODE closely following.

In summary, GN-SDE demonstrates consistent strength across datasets, highlighting its robustness and aptitude in handling out-of-distribution data with confidence.

\subsection{Spatio-Temporal}
\begin{figure}[ht]
\begin{adjustwidth}{-0cm}{-0cm}  

  \centering
  \begin{subfigure}{0.33\linewidth}
    \includegraphics[height=0.8\linewidth, width=\linewidth]{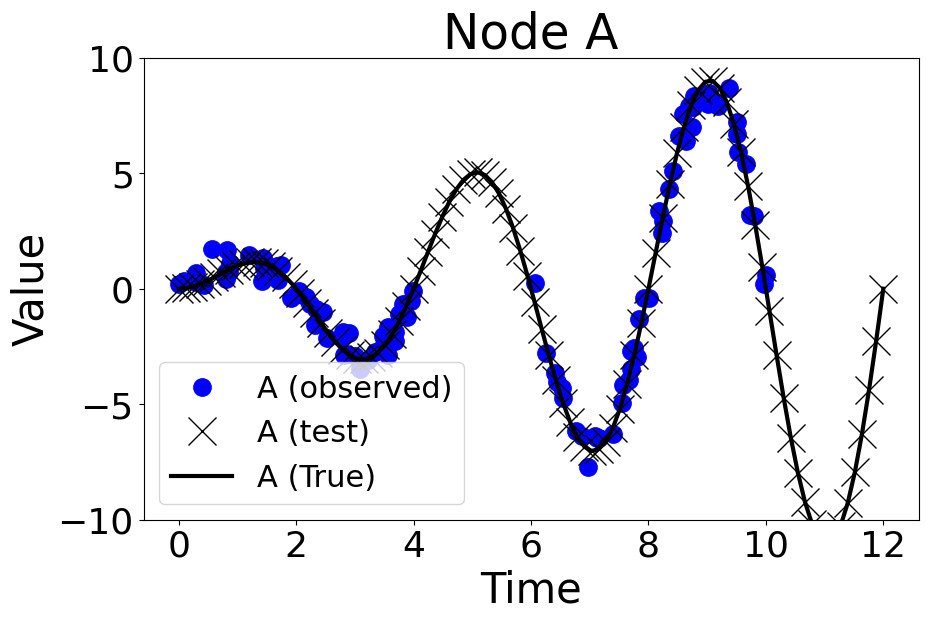}
  \end{subfigure}%
  \begin{subfigure}{0.33\linewidth}
    \includegraphics[height=0.8\linewidth, width=\linewidth]{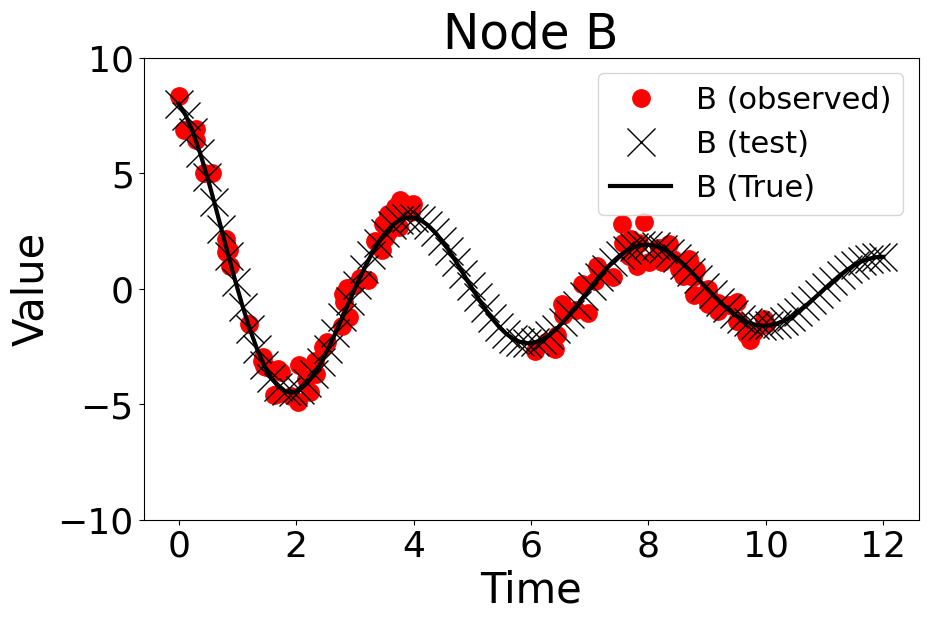}
  \end{subfigure}%
  \begin{subfigure}{0.33\linewidth}
    \includegraphics[height=0.8\linewidth, width=\linewidth]{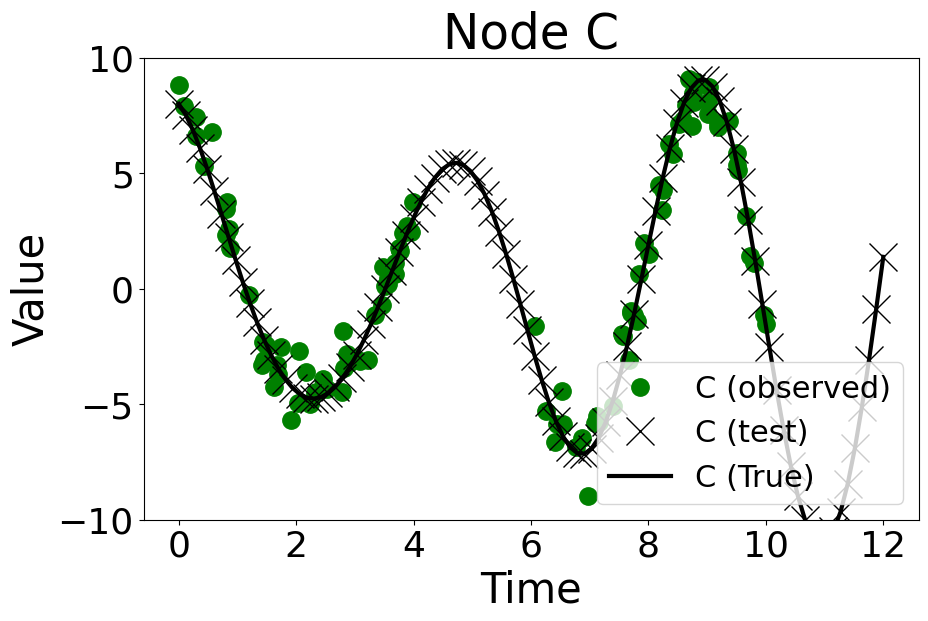}
  \end{subfigure}
  
  \begin{subfigure}{0.33\linewidth}
            \includegraphics[height=0.8\linewidth, width=\linewidth]{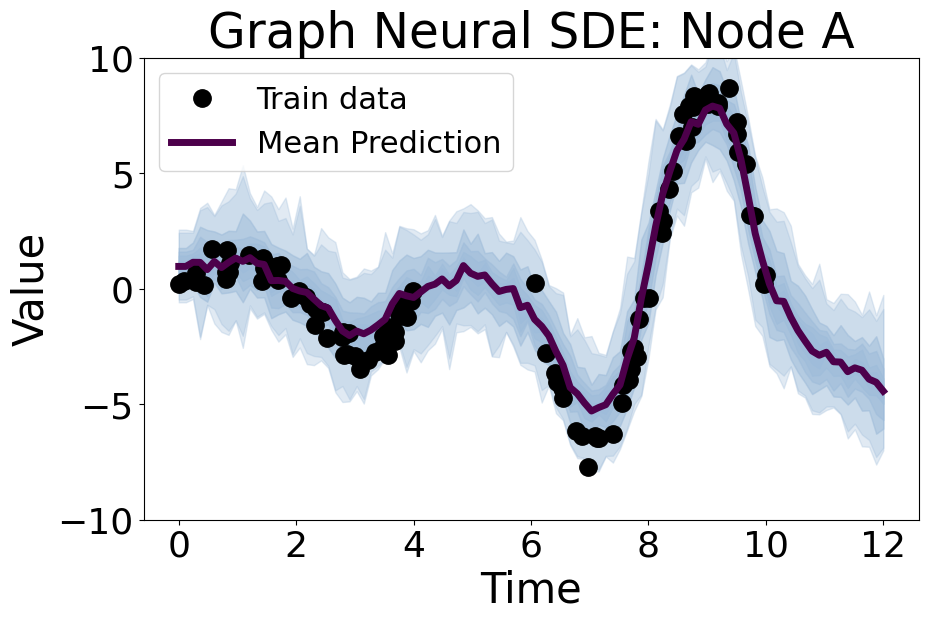}
  \end{subfigure}%
  \begin{subfigure}{0.33\linewidth}
    \includegraphics[height=0.8\linewidth, width=\linewidth]{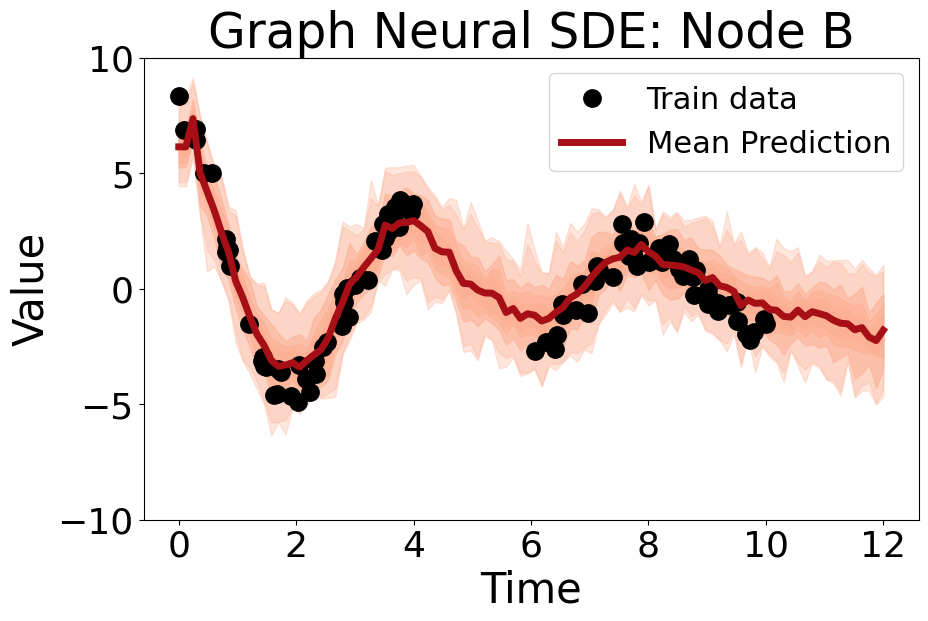}
  \end{subfigure}%
  \begin{subfigure}{0.33\linewidth}
    \includegraphics[height=0.8\linewidth, width=\linewidth]{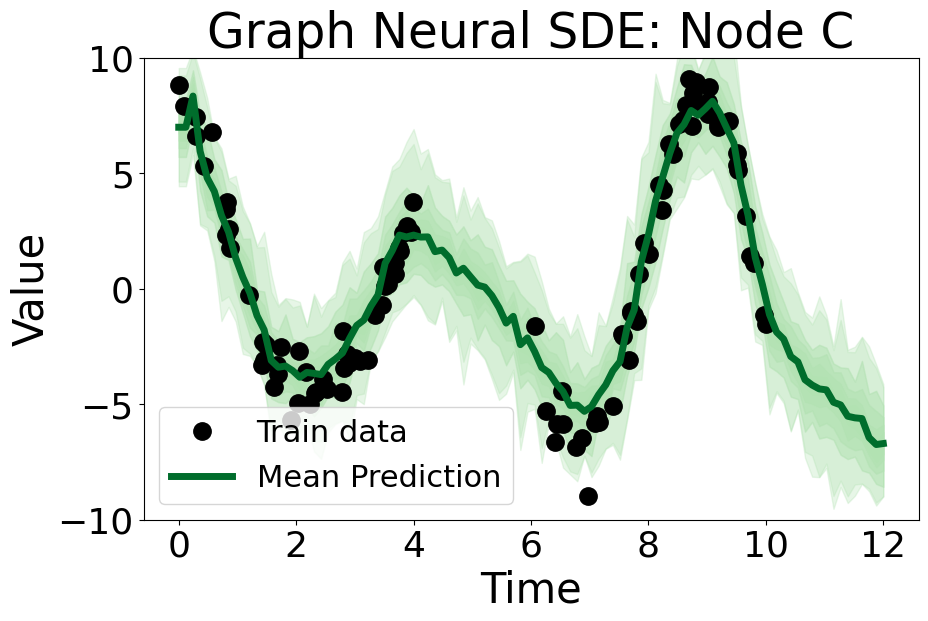}
  \end{subfigure}
  \end{adjustwidth}

  \caption{Comparison of the true and predicted values for the three-node regression problem. The first row shows the true values of nodes A, B, and C over time. The second row  presents the predictions made by the Graph Neural SDE model for nodes A, B, and C. The shaded regions represent the model's uncertainty quantification, demonstrating an increase in uncertainty during the interpolation and extrapolation phases.}

  \label{fig:SDE_vs_Toy_regression}
\end{figure}

Consider a three-node regression problem with nodes A, B, and C, where the goal is to predict their regression values at different time points. As shown in Figure \ref{fig:SDE_vs_Toy_regression}, the observations are irregularly sampled. The graph structure used has the node C is connected to nodes A and B, but nodes A and B are not directly linked. The figure also shows the training and testing data points, illustrating both interpolation (between time 4 and 6) and extrapolation (from time 10 to 12). The true distribution for each node is as follows:

\begin{equation*}
\begin{aligned}
A(t) &= t \cdot \sin\left(\frac{\pi t}{2}\right) + \epsilon_A, \quad
B(t) = \frac{4}{\frac{t}{5} + 0.5} \cdot \cos\left(\frac{\pi t}{2}\right) + \epsilon_B, \quad
C(t) = A(t) + B(t) + \epsilon_C.
\end{aligned}
\end{equation*}

where $\epsilon_A, \epsilon_B, \epsilon_C \sim \mathcal{N}(0, 0.5^2)$. It is important to note that while nodes A and B are independent, node C is a function of both A and B, with an added Gaussian noise component.
\\\\
Our goal is to integrate uncertainty into predictions during testing. Both Neural ODE and GCN models are deterministic, lacking a direct uncertainty measure for regression as entropy offers in classification. To introduce uncertainty, we use Monte Carlo Dropout during training and testing for these models. Under certain conditions, networks with dropout can emulate a Gaussian Process \citep{gal2016dropout}. This allows us to estimate a model's uncertainty by predicting both a mean and variance, using variance as the uncertainty measure. Predictions are made only if the variance meets a set threshold. The Bayesian GCN and Ensemble methods inherently quantify uncertainty, with the Ensemble calculating regression prediction variances and Bayesian GCN using Monte Carlo Sampling, aligning with our GN-SDE approach.

\begin{table}[ht]
    \centering
    \begin{adjustbox}{center}
    \begin{tabular}{l l cccccccc}
    \toprule
    \multirow{2}{*}{\underline{Metric}} & \multirow{2}{*}{\underline{Models}} & \multicolumn{6}{c}{Variance Thresholds} \\
    \cmidrule(r){3-8}
    & & 3 & 2.5 & 2 & 1.5 & 1.0 & 0.5 \\ \midrule
    \multirow{5}{*}{MAE} 
    & Dropout GCN & 13.35 & 13.06 & 13.03 & 13.37 & 12.24 & 5.631 \\
    & \textbf{GN-SDE (ours)} & 12.06 & 11.36 & 10.46 & 9.734 & 2.406 & - \\
    & Dropout GN-ODE & 14.52 & 14.52 & 14.52 & 14.49 & 13.61 & - \\
    & Bayesian GCN & 10.99 & 11.02 & 12.16 & 9.121 & 4.253 & 4.260 \\
    & Ensemble GCN & \textbf{2.665} & \textbf{2.678} & \textbf{2.737} & \textbf{2.715} & \textbf{2.615} & \textbf{2.736} \\ \midrule
    \multirow{5}{*}{MAPE} 
    & Dropout GCN & 9.340 & 9.390 & \textbf{9.462} & 14.08 & \textbf{8.941} & 3.060 \\
    & \textbf{GN-SDE (ours)} & \textbf{6.825} & \textbf{6.321} & 13.08 & \textbf{3.380} & 13.55 & - \\
    & Dropout GN-ODE & 8.909 & 8.909 & 8.909 & 10.67 & 9.956 & - \\
    & Bayesian GCN & 7.503 & 7.588 & 9.413 & 7.696 & 7.235 & 2.731 \\
    & Ensemble GCN & 15.44 & 141.20 & 25.15 & 54.49 & 18.31 & 8.612 \\ \midrule
    \multirow{5}{*}{MSE} 
    & Dropout GCN & 13.35 & 13.06 & 13.03 & 13.37 & 12.24 & 5.631 \\
    & \textbf{GN-SDE (ours)} & \textbf{12.06} & \textbf{11.36} & \textbf{10.46} & \textbf{9.734} & \textbf{2.41} & - \\
    & Dropout GN-ODE & 14.52 & 14.52 & 14.52 & 14.49 & 13.61 & - \\
    & Bayesian GCN & 10.99 & 11.02 & 12.16 & 9.121 & 4.253 & 4.260 \\
    & Ensemble GCN & 12.73 & 12.72 & 13.27 & 13.28 & 12.55 & 13.47 \\ \midrule
    \multirow{5}{*}{NLL} 
    & Dropout GCN & 22.55 & 25.07 & 25.29 & 26.74 & 26.66 & 19.81 \\
    & \textbf{GN-SDE (ours)} & \textbf{8.969} & \textbf{9.032} & \textbf{9.530} & \textbf{10.30} & 8.44 & - \\
    & Dropout GN-ODE & 40.19 & 40.19 & 40.19 & 42.04 & 43.97 & - \\
    & Bayesian GCN & 13.28 & 13.35 & 12.28 & 12.01 & \textbf{6.958} & 9.999 \\
    & Ensemble GCN & 51.38 & 85.56 & 81.54 & 89.95 & 104.68 & 205.54 \\ \bottomrule
    \end{tabular}
    \end{adjustbox}
    \caption{Performance comparison of Dropout GCN, GN-SDE, Dropout GN-ODE, and Bayesian GCN models on the three-node regression problem across different variance thresholds. Bold values indicate superior performance by the GN-SDE model.}
    \label{table:toy_regression}
\end{table}

Table \ref{table:toy_regression} shows the GN-SDE model outperforming the Dropout GCN and Dropout GN-ODE models across all variance thresholds. The GN-SDE model frequently achieves the lowest Mean Squared Error (MSE) and Negative Log-Likelihood (NLL) across the higher variance thresholds. However, its performance in Mean Absolute Percentage Error (MAPE) demonstrates some variability.

For the NLL metric, calculated under the Gaussian-distributed predictions assumption

\begin{equation*}
NLL = \frac{1}{N}\sum_{i = 1}^{N} \left( \frac{1}{2} \log(2\pi\sigma_i^2) + \frac{(y_i - \mu_i)^2}{2\sigma_i^2} \right),
\end{equation*}

where $\sigma^2$ is the predicted variance, $\mu$ is the observations, and  $\hat{\mu}$ is the predicted mean. 

As the variance threshold decreases, the Bayesian GCN becomes more competitive, even outperforming the GN-SDE in MAE at the 0.5 threshold. The Ensemble GCN's performance fluctuates, but it remains competitive. GN-SDE's MAE and MSE decline with decreasing variance, highlighting its capability to filter out uncertain predictions. At the 0.5 threshold, only the Dropout GCN and Bayesian GCN perform, with the latter showing remarkable MAE and NLL results.

In summary, GN-SDE is a standout in the three-node regression task, often surpassing peers across thresholds. The Bayesian GCN excels at lower variance thresholds. Dropout GCN holds its own at the strictest threshold, while Dropout GN-ODE struggles at higher variances, suggesting limitations in uncertainty assessment. The Ensemble GCN is consistent but sometimes underperforms in the NLL metric.

To evaluate the performance of these models on a real-world dataset, MET-LA, please refer to Appendix \ref{A:Meter-LA-resutls}.

\section{Related Work}

Uncertainty quantification in GNNs has been relatively less explored compared to traditional neural networks. Among the limited research in this domain, we have benchmarked our work against key contributions such as the Bayesian Graph Neural Network \cite{hasanzadeh2020bayesian} and robust ensemble methods \citep{lin2022robust}. While there has been significant research into Gaussian Processes on graphs \citep{borovitskiy2021matern, lawrence2004semi, perez2013gaussian}, we did not include these in our evaluations.

The use of differential equations on graphs is a growing research area, with most work focused on Graph Neural ODEs \citep{poli2019graph} and Graph Control Differential Equations \citep{choi2022graph}. Graph Neural ODEs have been applied to dynamic graph classification \citep{jin2022multivariate, shi2023towards}, traffic forecasting \citep{choi2022graph, liu2023graph}, and protein interface prediction  \citep{tanfuzzy3dvecnet}. Extensions include second-order and higher-order Graph ODEs \citep{luo2023hope, zhang2022improving}.

While SDEs provide a promising avenue for better uncertainty quantification in differential equations, their application in graphs has been limited. Notably, they are employed in Graph Diffusion models primarily for denoising purposes \citep{huang2022graphgdp, luo2022fast, jo2022score}.

In this landscape, the paper `BroNet' \citep{bishnoi2023graph} stands out for its claim to develop Graph Neural SDEs. Yet, its technique, which uses a Graph Neural Network to learn an SDE's scalar parameter before integrating it, fails to model the graph as an SDE, setting it apart from our approach.

In essence, the field of uncertainty quantification in GNNs remains relatively unexplored. Our model presents a novel integration of SDE with GNN to address this gap. We hope that our contribution acts as a cornerstone for further developments in Graph Neural SDE-based uncertainty quantification.

\section{Conclusion}
This study introduces the \textbf{Graph Neural Stochastic Differential Equations } model, by using the Latent approach, designed for tasks like node, graph, and link prediction. Experimental results show that Latent Graph Neural SDEs consistently outperform models such as the Bayesian GCN, Ensemble GCN, and Graph Neural ODEs, excelling in metrics like uncertainty quantification and out-of-distribution detection. The superior accuracy of our model might stem from the intrinsic noise within its differential equations, similar to data augmentation in traditional machine learning. This noise may enhance model robustness and testing generalization, this hypothesis needs future exploration — perhaps by comparing a Graph Neural ODE trained on noise-augmented data and assessing the resulting accuracy differences. However, the advanced nature of Graph Neural SDEs requires increased computational resources, primarily due to the elevated integration costs of SDEs which make it more computationally expensive than the other models.

Moving forward, several intriguing avenues could be explored. One such avenue is the Bayesian Neural Network SDE that employs an Ornstein-Uhlenbeck prior for its weights, as described by \cite{xu2022infinitely}. This approach could be extended to graph SDEs, potentially leading to the development of Graph Bayesian Neural Network SDEs (Graph BNN-SDEs). Additionally, the advancements in Partial Differential Equations showcased by \cite{sun2020neupde} can be creatively adapted for graph contexts using a latent method similar to our approach, paving the way for Graph Stochastic Partial Differential Equations (Graph SPDEs). Moreover, integrating higher-order stochastic differential equations with the Latent Graph Neural SDE we introduced might be advantageous and could further enhance the modeling of complex systems.

\bibliographystyle{abbrvnat}
\bibliography{references.bib}

\appendix

\newpage

\section{What Are Graph Neural ODE and Graph Neural SDEs anyways?}
\label{A:RNN_REsNET}
Here, we demonstrate the equivalences or representations of Graph Neural ODEs and Graph Neural SDEs in relation to popular architectures.
\subsection{Graph Neural ODEs as Continuously-deep Graph Residual Neural Networks}
In this section, we present a straightforward extension of this idea to graph structures. Specifically, we illustrate that a Graph Neural ODE can be conceptualized as a continuous-depth version of a Graph Residual Network.

Considering the architecture of a residual graph network:

\begin{equation}
\label{eq:resNet2}
y_{j+1} = y_j + f_\mathcal{G}(j, y_j, \theta)
\end{equation}
where $f_\Theta(j, y_j, \mathcal{G})$ represents the $j$-th residual block, with the parameters from all blocks being collectively represented by $\Theta$.
\\
In contrast, let's look at the Graph Neural ODE (abbreviated as GN-ODE):

\begin{equation*}
\frac{dy}{dt}(t) = f_\mathcal{G}(j, y_j, \theta)
\end{equation*}
\\
Discretizing this GN-ODE using the explicit Euler method at uniformly spaced time intervals $t_j$ with a gap of $\Delta t$ gives:

\begin{equation*}
\frac{y(t_{j+1}) - y(t_j)}{\Delta t} \approx \frac{dy}{dt}(t_j) =  f_\mathcal{G}(j, y_j, \theta)
\end{equation*}
\\
Simplifying, we get:

\begin{equation*}
y(t_{j+1}) = y(t_j) + \Delta t \cdot  f_\mathcal{G}(j, y_j, \theta)
\end{equation*}

By integrating the factor of $\Delta t$ into $f_\mathcal{G}$, this equation naturally aligns with the formulation in Equation \ref{eq:resNet2}. Such a perspective underscores that neural ODEs can be seen as the continuous-time counterparts of residual networks.

This viewpoint casts the graph neural ODE as a continuously-deep Graph Residual Network. Here, the sequence of minor (residual) updates to its hidden states become both infinitesimally small and infinitely frequent. The end output is the cumulative effect of these continuous updates, mirroring the solution to the ODE from its initial state.

\subsection{Graph Neural SDEs as Continuously-Deep Recurrent Graph Neural Networks}
A well-established analogy exists between numerically discretized neural stochastic differential equations and the structures found in deep learning literature - particularly Recurrent Neural Networks (RNNs). In the case of Neural SDEs, the RNN's input can be interpreted as random noise, or Brownian motion, while its output corresponds to a generated sample.
\\\\
Consider an autonomous one-dimensional Itô Stochastic Differential Equation represented as:

\begin{equation*}
dy(t) = f(y(t)) dt + \sigma(y(t)) dw(t)
\end{equation*}

where $y(t)$, $f(y(t))$, $\sigma(y(t))$, and $w(t)$ belong to the set of real numbers, $\mathbb{R}$. The numerical Euler-Maruyama discretization of this SDE can be expressed as:
\begin{equation*}
\frac{y(t_{j+1}) - y(t_j)}{\Delta t} \approx f(y(t_j)) + \frac{\sigma(y(t_j))\Delta w_j}{\Delta t}
\end{equation*}
Which simplifies to:
\begin{equation*}
y_{j+1} = y_j + f(y_j)\Delta t + \sigma(y_j)\Delta w_j
\end{equation*}

Here, $\Delta t$ represents a fixed time step and $\Delta w_j$ is normally distributed with mean zero and variance $\Delta t$. This numerical discretization is reminiscent of an RNN with a specific form. Therefore, we can consider the Neural SDE as an consciously-deep RNN where the depth is defined by the number of discretization steps of the SDE solver.

\section{Extended Results}
\label{extended_static_results}

\subsection{Spatial-Temporal Images}
\begin{figure}[ht]
\begin{adjustwidth}{-0cm}{-0cm}  

  \centering
  \begin{subfigure}{0.33\linewidth}
    \includegraphics[width=\linewidth]{True_A.png}
  \end{subfigure}%
  \begin{subfigure}{0.33\linewidth}
    \includegraphics[width=\linewidth]{True_B.png}
  \end{subfigure}%
  \begin{subfigure}{0.33\linewidth}
    \includegraphics[width=\linewidth]{True_C.png}
  \end{subfigure}

  \begin{subfigure}{0.33\linewidth}
    \includegraphics[width=\linewidth]{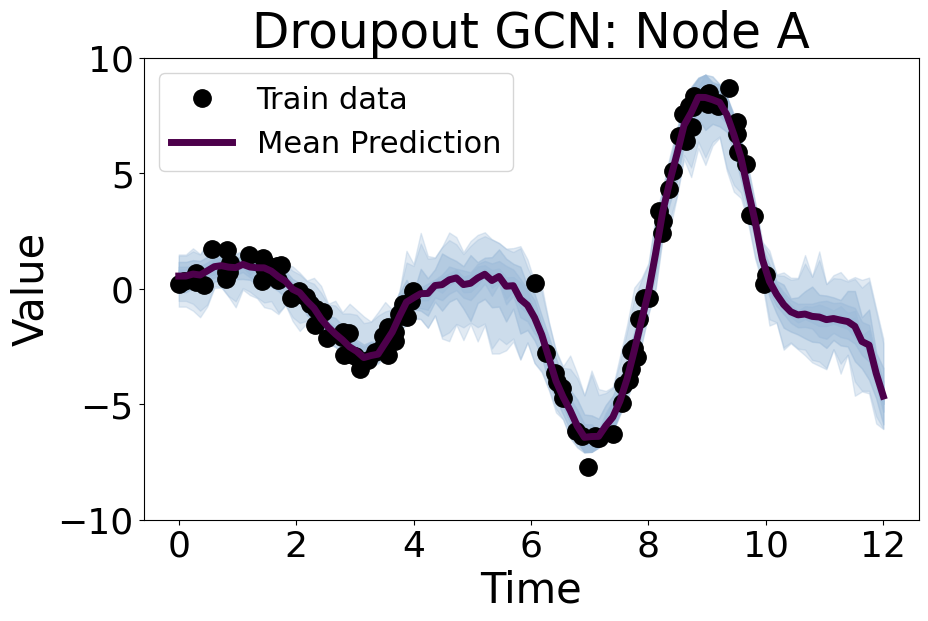}
  \end{subfigure}%
  \begin{subfigure}{0.33\linewidth}
    \includegraphics[width=\linewidth]{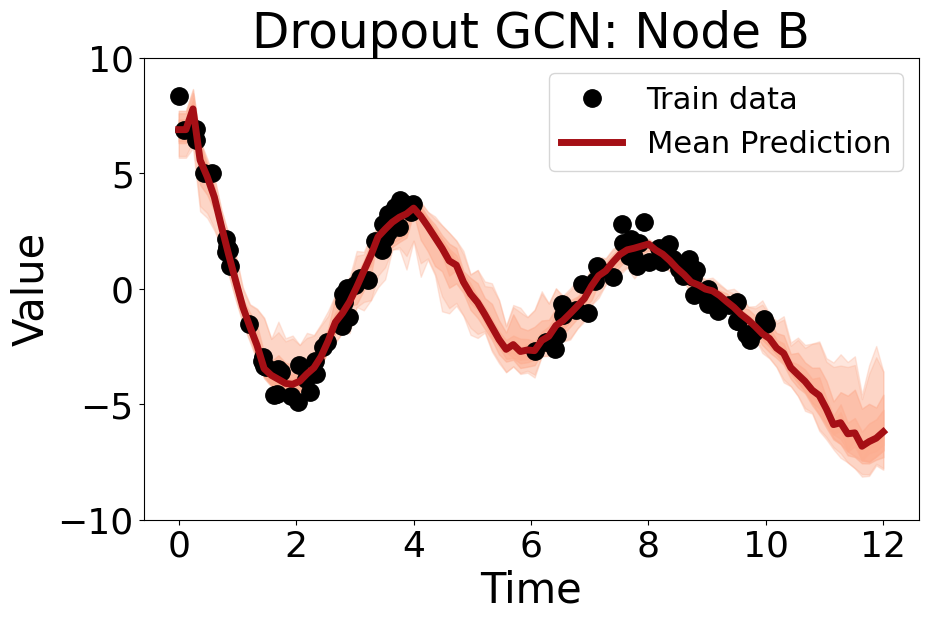}
  \end{subfigure}%
  \begin{subfigure}{0.33\linewidth}
    \includegraphics[width=\linewidth]{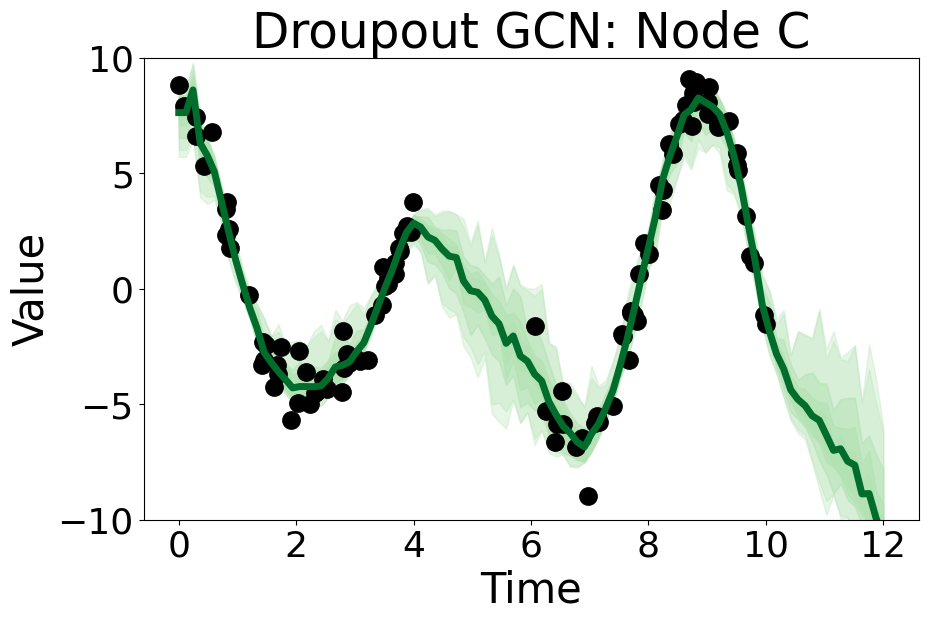}
  \end{subfigure}

  \begin{subfigure}{0.33\linewidth}
    \includegraphics[width=\linewidth]{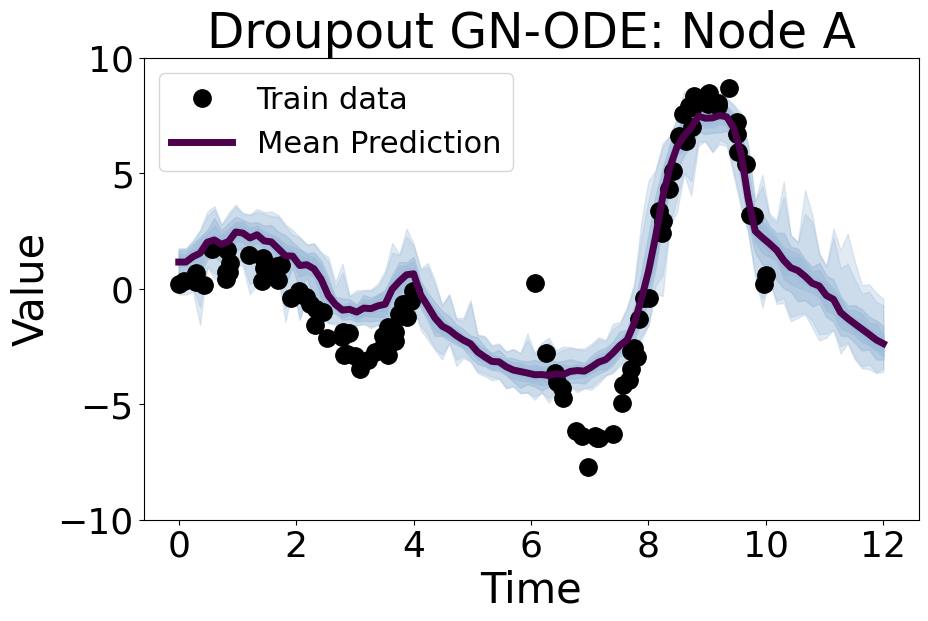}
  \end{subfigure}%
  \begin{subfigure}{0.33\linewidth}
    \includegraphics[width=\linewidth]{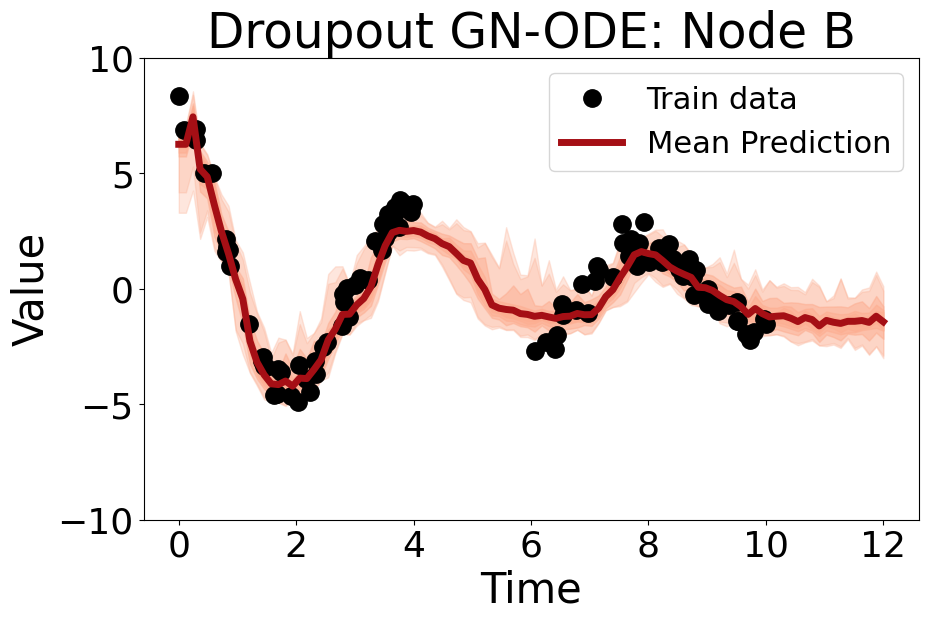}
  \end{subfigure}%
  \begin{subfigure}{0.33\linewidth}
    \includegraphics[width=\linewidth]{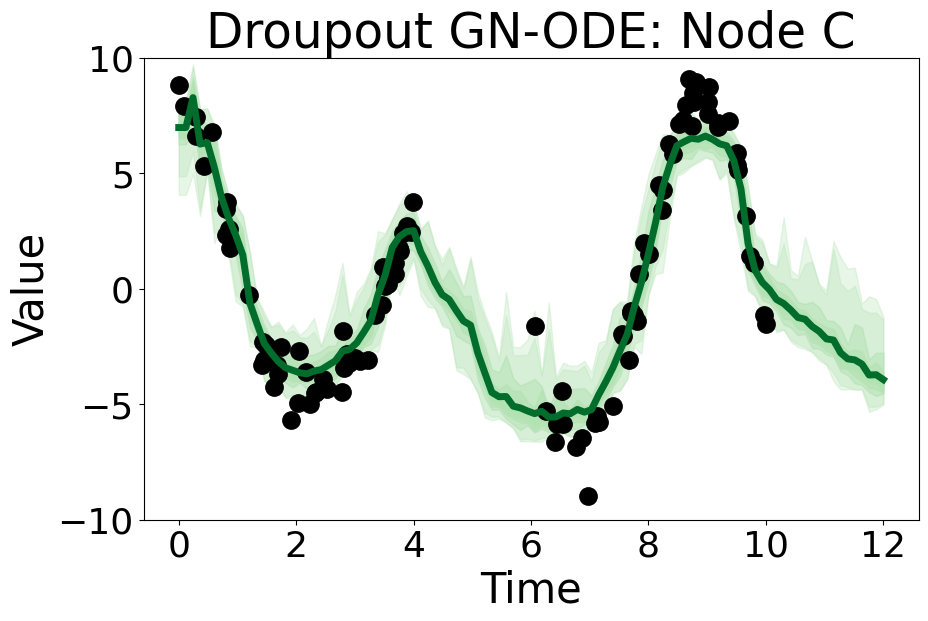}
  \end{subfigure}

  \begin{subfigure}{0.33\linewidth}
            \includegraphics[width=\linewidth]{SDE_A.png}
  \end{subfigure}%
  \begin{subfigure}{0.33\linewidth}
    \includegraphics[width=\linewidth]{SDE_B.png}
  \end{subfigure}%
  \begin{subfigure}{0.33\linewidth}
    \includegraphics[width=\linewidth]{SDE_C.png}
    
  \end{subfigure}
  \end{adjustwidth}
    \caption{The figure displays 12 images organized. Each column corresponds to one of the 3 nodes, while each row represents a different model or dataset. The top row illustrates the training and testing datasets for each node, in the context of node regression. The aim is to predict the regression value for each node. The second row presents results from the GCN, the third row showcases those from the Graph Neural ODE, and the bottom row depicts our model, the Graph Neural SDE.}
    \label{fig:sde_ode_gcd_dropout_9_fig}
\end{figure}

\subsection{Meter-LA Experiment Set Up}
\label{A:Meter-LA Set U}
In our experiments, we utilized Graph Attention Networks (GAT) to embed the input data, which consisted of the past six recordings. These inputs were embedded into 64-dimensional tensors. Subsequently, these embeddings were passed to our Latent Graph Stochastic Differential Equation (SDE) model. The Graph Latent SDE model employs a  GCN for the drift function, while the diffusion function is held constant at a value of 1. The hidden state of the model is of size 64, which is then passed to a GCN projection layer for prediction.
\\\\
The Graph Neural ODE model follows a similar structure but replaces the SDE with an ODE and omits the noise component. The GCN model also utilizes the same embedding and projection layers but bypasses the differential equations entirely.
\\\\
This setup allows for a fair comparison between the models, as they share the same embedding and projection layers, differing only in the differential equation component.

\subsubsection{Real World Datasets}
\label{A:Meter-LA-resutls}
Transitioning to real-world spatiotemporal datasets, we focus on METR-LA for traffic prediction, a typical time-series problem. The goal is to forecast future traffic metrics, such as speed or flow, over the next \(H\) steps based on previous \(M\) steps' traffic observations~\citep{li2018dcrnn_traffic}. This can be mathematically described as:

\begin{equation*}
\hat{v}_{t+1},\ldots,\hat{v}_{t+H} = \arg\max_{v_{t+1},\ldots,v_{t+H}} \log P(v_{t+1},\ldots,v_{t+H}|v_{t-M+1},\ldots,v_{t}),
\end{equation*}

where \(v_t \in \mathbb{R}^n\) is an observation vector at time \(t\) for \(n\) road segments~\citep{yu2017spatio}. The METR-LA dataset, collected from 207 highway loop detectors in LA County, contains four months of data from March to June 2012. Data was recorded every 5 minutes, and our experiments aim to predict an hour's worth of traffic speed (12 readings) based on the last hour's data. Please refer to appendix \ref{A:Meter-LA Set U} for experimental setup details.

\begin{table}[ht]
    \centering
    \begin{adjustbox}{center}
    \begin{tabular}{l l cccccc}
    \toprule
    \multirow{2}{*}{\underline{Metric}} & \multirow{2}{*}{\underline{Models}} & \multicolumn{5}{c}{Variance Thresholds} \\
    \cmidrule(r){3-7}
    & & 100 & 3 & 1 & 0.5 & 0.25 \\ \midrule
    \multirow{5}{*}{MAE} 
    & Dropout GAT-GCN & 14.30 & 14.01 & 9.38 & 5.25 & 5.22 \\
    & \textbf{GN-SDE (ours)} & 14.13 & 12.42 & 10.14 & 5.40 & \textbf{4.82} \\
    & Dropout GN-ODE & 15.21 & 15.53 & 12.36 & 9.37 & 5.58 \\
    & Bayesian GCN & 15.0 & 13.9 & 9.4 & 5.3 & 5.2 \\
    & Ensemble GCN & \textbf{9.835} & \textbf{6.665} & \textbf{2.615} & \textbf{2.736}& - \\ \midrule
    \multirow{5}{*}{MAPE} 
    & Dropout GAT-GCN & \textbf{10.21} & 10.59 & 20.42 & 21.01 & 20.87 \\
    & \textbf{GN-SDE (ours)} & 10.63 & \textbf{9.66} & \textbf{18.74} & \textbf{18.79} & \textbf{13.25} \\
    & Dropout GN-ODE & 10.27 & 13.92 & 21.86 & 17.14 & 13.70 \\
    & Bayesian GCN & 10.5 & 10.6 & 20.5 & 20.9 & 20.8 \\
    & Ensemble GCN & 18.15 & 15.44 & 18.31 & 8.61 & - \\ \midrule
    \multirow{5}{*}{RMSE} 
    & Dropout GAT-GCN & 18.58 & 16.39 & 9.58 & 5.26 & 5.22 \\
    & \textbf{GN-SDE (ours)} & \textbf{15.38} & \textbf{13.17} & \textbf{10.12} & \textbf{5.22} & \textbf{4.78} \\
    & Dropout GN-ODE & 19.73 & 17.32 & 16.68 & 12.87 & 7.33 \\
    & Bayesian GCN & 19.0 & 16.5 & 9.6 & 5.3 & 5.2 \\
    & Ensemble GCN & 19.74 & 12.73 & 7.55 & 6.47 & - \\ \midrule
    \multirow{5}{*}{RNLL} 
    & Dropout GAT-GCN & 7.03 & 13.28 & 15.12 & 11.56 & 14.52 \\
    & \textbf{GN-SDE (ours)} & \textbf{6.56} & \textbf{11.57} & \textbf{13.54} & \textbf{13.92} & \textbf{13.88} \\
    & Dropout GN-ODE & 32.33 & 75.10 & 31.03 & 92.42 & 213.09 \\
    & Bayesian GCN & 7.2 & 13.4 & 15.3 & 12.0 & 14.6 \\
    & Ensemble GCN & 39.63 & 51.38 & 104.68 & 205.54 & - \\ \bottomrule
    \end{tabular}
    \end{adjustbox}
    \caption{Comparative performance of various models on the METER-LA Dataset across different variance thresholds. Bold values indicate superior performance by the model.}
    \label{table:toy_regression2}
\end{table}

In Table \ref{table:toy_regression2}, the GN-SDE model consistently surpasses other contenders across all variance thresholds, notably securing the lowest MAE, MAPE, RMSE, and RNLL scores. This emphasizes GN-SDE's stellar accuracy and unmatched uncertainty estimation abilities. Even with stricter variance thresholds of 0.5 and 0.25, GN-SDE remains at the forefront, signifying its effective prediction alignment with observed data and adeptness at uncertainty quantification.

Specifically, the Dropout GN-ODE struggles with the RNLL metric, hinting at a subpar fit and diminished uncertainty evaluation capabilities. This shortcoming becomes glaringly apparent at the 0.25 variance threshold, where its RNLL significantly eclipses that of its peers. Bayesian GCN deserves mention for its resilient performance, often ranking a close second to the GN-SDE, underlining its robust modeling and uncertainty estimation prowess. Meanwhile, Ensemble GCN displays intermittent strengths in the MAE at lower variances but demonstrates inconsistency, particularly with fluctuating RNLL scores at the 100 and 3 variance thresholds.

In summation, while GN-SDE dominates in its robustness across varying thresholds, Bayesian GCN also emerges as a formidable model. Dropout GAT-GCN and Dropout GN-ODE especially grapple with the RNLL metric, and Ensemble GCN exhibits variable results.

\end{document}